\documentclass{interact}

\usepackage[colorinlistoftodos]{todonotes}
\usepackage{tikz}
\usepackage{soul}
\usetikzlibrary{calc}
\usetikzlibrary{positioning,shapes,arrows.meta}
\usepackage{graphicx}
\usepackage{subcaption}
\usepackage{float} 
\usepackage{geometry}
\usepackage{fix-cm}
\usepackage{epstopdf}

\usepackage{natbib}
\bibpunct[, ]{(}{)}{;}{a}{}{,}
\renewcommand\bibfont{\fontsize{10}{12}\selectfont}

\theoremstyle{plain}

\theoremstyle{definition}

\theoremstyle{remark}

\begin{document}


\title{Water Mapping and Change Detection Using Time Series Derived from the Continuous Monitoring of Land Disturbance Algorithm}

\author{
\name{Huong Pham\textsuperscript{a}, Samuel Cheng\textsuperscript{a}, Tao Hu\textsuperscript{b},
and Chengbin Deng\textsuperscript{c}}
\affil{\textsuperscript{a}School of Electrical and Computer Engineering, University of Oklahoma, OK, USA; \textsuperscript{b}Department of Geography, Oklahoma State University, OK, USA; \textsuperscript{c}Center for Spatial Analysis, Department of Geography, University of Oklahoma, OK, USA}
}

\maketitle

\begin{abstract}
Given the growing environmental challenges, accurate monitoring and prediction of changes in water bodies are essential for sustainable management and conservation. The Continuous Monitoring of Land Disturbance (COLD) algorithm provides a valuable tool for real-time analysis of land changes, such as deforestation, urban expansion, agricultural activities, and natural disasters. This capability enables timely interventions and more informed decision-making. This paper assesses the algorithm’s effectiveness to estimate water bodies and track pixel-level water trends over time. Our findings indicate that COLD-derived data can reliably estimate estimate water frequency during stable periods and delineate water bodies. Furthermore, it enables the evaluation of trends in water areas after disturbances, allowing for the determination of whether water frequency increases, decreases, or remains constant.

\end{abstract}


\section{Introduction}
Detecting surface water in aquatic ecosystems is essential for environmental monitoring, water resource management, and ecosystem conservation, particularly in communities where water resources are closely linked to recreational and economic activities. The decline in water levels in many lakes has raised significant concerns, highlighting the need for accurate detection methods to assess water availability, understand hydrological cycles, and manage droughts and floods. Remote sensing offers a valuable tool for measuring the spatial variability of surface water in these ecosystems. Advances in remote sensing technologies and satellite imagery have significantly improved our ability to monitor water bodies, providing critical data for hydrological analysis and identifying the drivers of water level fluctuations \citep{duanRecentAdvancementRemote2021}.

The Continuous Monitoring of Land Disturbance (COLD) algorithm, developed using Landsat time series data, offers an effective approach for detecting various types of land disturbances as new imagery becomes available \citep{zhuContinuousMonitoringLand2020}. Additionally, it suports the generation of historical maps of these disturbances. Through experimentation with different data inputs and time series analysis techniques, the algorithm has been refined to enhance detection performance. Key improvements include the use of surface reflectance instead of Top-of-Atmosphere (TOA) reflectance, the integration of multiple spectral bands, and the efficient removal of outliers based on model predictions. Importantly, the algorithm can distinguish between actual land disturbances and changes resulting from vegetation regrowth, improving the accuracy of land disturbance mapping. This capability lays a robust foundation for investigating the algorithm's potential to estimate water frequency, defined as the proportion of observations classified as water relative to the total number of stable observations, where no disturbances occur. This enables the delineation of water bodies and the analysis of water trends at the pixel level following disturbance events.

Generating land cover classifications from remotely sensed data is a relatively straightforward process, but achieving high accuracy presents challenges. To enhance classification accuracy, many products incorporate multitemporal images into their algorithms \citep{melicharRandomForestClassification2023, liuInvestigationCapabilityMultitemporal2021, shettyAssessingEffectTraining2021}. However, automating the collection of multitemporal images can introduce several potential issues. One major challenge is ensuring data consistency across images. Variations in atmospheric conditions during different image acquisition dates can affect spectral signatures, making it difficult for algorithms to maintain consistent classification over time. For instance, cloud-free images are essential for accurately classifying each pixel across multiple observations. This requirement is often unattainable, particularly for sensors with low temporal frequency, such as Landsat. As a result, acquiring suitable images may take several years, leading to the production of Landsat-based land cover maps at five- or ten-year intervals \citep{zhuContinuousChangeDetection2014}.

Additionally, temporal variability in images, such as phenological changes in vegetation or land use changes due to human activities, can lead to misclassification if the algorithm does not account for these temporal dynamics \citep{wangMountainVegetationClassification2024}. This issue is particularly pronounced when images from long intervals are used or when mapping areas that undergo frequent changes. Consequently, land cover maps produced using conventional multitemporal methods may be unreliable for identifying land cover changes, as classification errors can be misinterpreted as actual changes. This problem is further exacerbated in smaller areas where subtle changes are more likely to occur.

To address these challenges, using data derived from Continuous Monitoring of Land Disturbance (COLD) algorithm could be a promising approach for classifying water bodies, as it (1) accounts for the temporal dynamics of land cover changes, (2) is robust against data inconsistencies, and (3) reliably detects water change trends. The main contributions of this study are as follows:
\begin{enumerate}
  \item The proposed classification and regression models can effectively identify the water frequency of various areas.
  \item Estimate the changing in water frequency at the break time between two stable periods.
\end{enumerate}

\section{Related work}
Various methods for extracting water surfaces using remote sensing data have been developed, including both machine learning and traditional approaches \citep{gharbiaDeepLearningAutomatic2023}. Machine learning methods require large datasets and careful sample selection to ensure generalization across different regions. Traditional algorithms, in contrast, often rely on spectral differences between water and non-water areas, using multiple bands for effective extraction. A notable example is the water index method, which combines spectral bands and remote sensing indices to distinguish water bodies. This method has demonstrated high performance in studies and offers advantages such as simplicity, robustness, and ease of deployment across large, diverse areas, enabling quick and efficient extraction \citep{wangTrackingAnnualChanges2020}.

Deep learning, with the invention of convolutional neural networks (CNNs), has been particularly useful for segmentation tasks in remotely sensed image analysis. These models can automatically extract both low- and high-level features from images, enabling the classification of water bodies with high accuracy, even in challenging environments \citep{changMultiScaleAttentionNetwork2024, mullenUsingHighResolutionSatellite2023, nasirDeepLearningDetection2023, liuNovelDeepLearning2024, erdemEnsembleDeepLearning2021, gharbiaDeepLearningAutomatic2023}. For example, the CNN-based model U-Net has been widely used in segmentation tasks \citep{caoWaterBodyExtraction2024, gharbiaDeepLearningAutomatic2023}. An ensemble model, WaterNet, combines various U-Net architectures with the cGAN-based Pix2Pix to improve water body segmentation accuracy \citep{erdemEnsembleDeepLearning2021}. Similarly, Wang \citep{wangSAMRSScalingupRemote2023} investigates the SER34AUnet model, an enhancement of the U-Net architecture specifically tailored for water extraction in cold and arid regions. While deep learning models are powerful for tasks such as water body extraction, they are heavily data-dependent and require large amounts of labeled data for training. Acquiring large, continuous, high-quality labeled datasets can be challenging and time-consuming \citep{gharbiaDeepLearningAutomatic2023}. Due to the difficulty of acquiring high-quality training datasets, these models are prone to overfitting, especially when the dataset is small or not representative of the broader context, leading to poor generalization on unseen data \citep{liuNovelDeepLearning2024}.

Recent advances in visual foundation models like the Segment Anything Model (SAM) \citep{kirillovSegmentAnything2023c} and Contrastive Language-Image Pre-training (CLIP) \citep{radfordLearningTransferableVisual2021a} show strong potential in various computational tasks. SAM, while effective at distinguishing features such as water from shadows in urban environments, faces challenges in accurately segmenting water bodies during low water levels and seasonal changes \citep{ozdemirExtractionWaterBodies2024}. This often leads to under-segmentation of shorelines, especially in low-water regions. SAM also struggles with high color contrast in complex environments, causing inaccurate water segmentation and misclassifications \citep{nasirDeepLearningDetection2023}. Since neither SAM nor CLIP were trained on remote sensing imagery, their performance in such applications is limited. Fine-tuning with more representative datasets could improve accuracy but requires additional resources \citep{ozdemirExtractionWaterBodies2024}.

These models face challenges, such as the need for cloud-free, high-resolution images, which are difficult to obtain consistently for long-term water change detection \citep{yueFullyAutomaticHighaccuracy2023, caoWaterBodyExtraction2024, ozdemirExtractionWaterBodies2024}. Many methods rely on individual images for segmentation models, making them sensitive to cloud cover \citep{yuliantoEvaluationThresholdImproved2022}. However, the Continuous Change Detection and Classification (CCDC) model by \citep{zhuContinuousChangeDetection2014}, which uses all Landsat data, addresses this by building a dynamic time series model that detects various land cover types, including water. Its updated version, Continuous Monitoring of Land Disturbance (COLD), enhances small-change detection, automates processing, ensures model stability, and handles cloud and snow effects \citep{zhuContinuousMonitoringLand2020}. Our proposed model leverages this data to estimate water indices, classify water bodies, and monitor water trends across regions.

\section{Study site and datasets}

This section examines the use of Landsat data to monitor surface water changes, essential for understanding ecological and hydrological processes. We discuss the selection of imagery from the United States Geological Survey from 1984 to 2001, focusing on low-cloud-cover images and specific water bodies. We also highlight the advantages of Landsat’s free availability and moderate resolution for large-scale environmental monitoring.

\subsection{Landsat data}
Monitoring surface water is crucial for understanding ecological and hydrological processes. Recent advancements in satellite-based optical remote sensing have significantly enhanced surface water detection \citep{huangDetectingExtractingMonitoring2018}. A common method involves using Landsat remote sensing data and water indices to track changes in water bodies throughout the year over large areas \citep{yueFullyAutomaticHighaccuracy2023, fengLongtermDenseLandsat2022}.

For this study, we obtained Landsat imagery from January to December (1984–2001) from the United States Geological Survey, covering both wet and dry seasons. Landsat images were chosen for their free availability, moderate resolution, and suitability for our study scale and computational resources. Images were selected based on quality, scene availability, and a cloud cover threshold of less than 5\% annually. Water bodies such as Lake Stanley Draper, Carl Blackwell Lake, Shawnee Reservoir, and Sooner Lake were included. For most years, data from more than half of the months were available. One image per month was selected when possible; months with unsuitable images or excessive cloud cover were excluded. The final dataset, encompassing seven regions, was consolidated for model training and validation, ensuring accurate water body identification and trend detection.

\subsection{COntinuous monitoring of Land Disturbance}

The COLD data aligns with the tiling system used by the Landsat Analysis Ready Data (ARD) \citep{dwyerAnalysisReadyData2018} across the Continental United States (CONUS), which consists of 427 mapped tiles. Each tile references the Albers Equal Area Conic projection and is based on the WGS84 datum. The tiles are uniformly sized at 5,000 by 5,000 pixels, with each pixel representing a 30-meter square, covering an area of 150 by 150 kilometers per tile. Tile identification is facilitated through horizontal ('h') and vertical ('v') coordinates.

Our specific focus is on tile h12v16, located in Oklahoma, which was selected for its availability of COLD data. The COLD data is aggregated into a 5000x5000 matrix file, with the dataset covering the time frame from 1994 to 2001. Each pixel contains one or more stable time points based on multiple coefficients.

\section{Methods}
\begin{figure}[htbp]
    \centering
    \includegraphics[width=\linewidth]{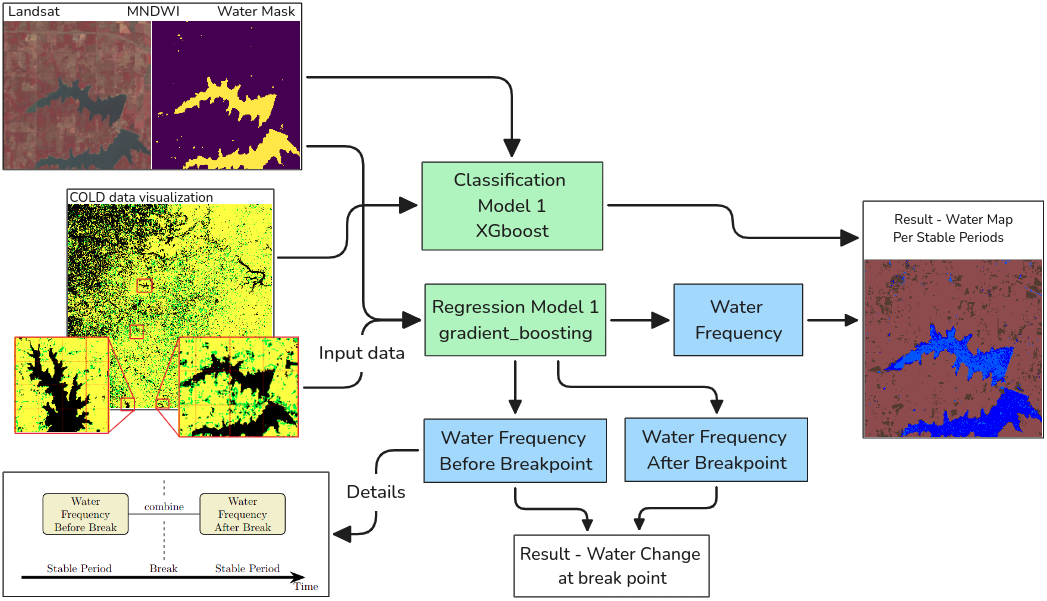}
    \caption{Methodology to predict the water mapping and water change from COLD data.}
    \label{fig:general}
\end{figure}

In this section, we describe the methodology used to build a dataset for training machine learning models to characterize water surfaces from 1984 to 2001. The Modified Normalized Difference Water Index (MNDWI) was employed to extract water body maps due to its simplicity and proven performance \citep{yueFullyAutomaticHighaccuracy2023, yuliantoEvaluationThresholdImproved2022, bijeeshGenericFrameworkChange2021}. Based on the water index, water bodies were delineated and used as inputs for the machine learning models. Combined with COLD data, we developed various models to estimate the average water index or water frequency. This approach effectively distinguishes between water and non-water bodies while identifying pixel-wise changes in water trends. The general approach is depicted in Figure \ref{fig:general}.

\subsection{Water bodies extraction}
Water indices, derived from two or more spectral bands, provide a straightforward and effective method for water extraction \citep{yueFullyAutomaticHighaccuracy2023, yuliantoEvaluationThresholdImproved2022, huangDetectingExtractingMonitoring2018}. Various indices can estimate surface water areas or flood sizes. The MNDWI is effective for this task using the shortwave infrared band (SWIR) for better detection of minor water features and being less affected by sediment and other water constituents, making it more reliable \citep{xuModificationNormalisedDifference2006}. These indices have proven effective in multiple studies \citep{gongApplicabilitySurfaceWater2020, minhtrinhApplicationMNDWIIndex2024}, providing a solid foundation for building the input data for training models to utilize COLD data for detecting water surfaces and their changes over time. 

To build the dataset for training machine learning models to characterize water surfaces from 1984 to 2001, the MNDWI was used to extract water features. It was widely adopted for its efficiency and reliability, enabled the extraction of water and land masks from lake images \citep{chisadzaDetectingLandSurface2022, gongApplicabilitySurfaceWater2020}. A manual threshold of zero, chosen for its simplicity and accuracy, classified pixels as water or non-water. This threshold served as the ground truth to assess the correlation between COLD data and water bodies. Using Landsat 5 data, the MNDWI was calculated from bands 2 and 5 \citep{mahuaANALYSISCOASTLINECHANGES2023}.

\subsection{Water and non-water bodies classifier using COLD-derived data}

The water indices extracted monthly from the stable periods of COLD-derived data within a given year are used to calculate the water frequency or the average water indices over multiple years. These indices represent the water index for each pixel in the COLD-derived data, corresponding to different stable periods. Water frequency is defined as the proportion of times a pixel is classified as water during a stable period. It is calculated by averaging the water observations from the available months in each year within that period. The frequency value ranges from 0 to 1, where 0 indicates no water observed, and 1 signifies that the pixel was consistently classified as water, as determined by the MNDWI.

Each pixel has 56 coefficients corresponding to different stable periods from the COLD-derived data, which are used as input features to estimate water frequency within these intervals. To mitigate potential errors arising from uncertainties in image processing, pixels with a water frequency of less than 0.25 are classified as non-water (land) areas, while pixels with a water frequency greater than 0.25 are categorized as water bodies \citep{fengLongtermDenseLandsat2022, yueFullyAutomaticHighaccuracy2023}.

Leveraging the water index as a benchmark and employing COLD-derived data as input features, we developed regression and classification models to estimate water frequency and distinguish between water and non-water bodies for each pixel. These models include gradient boosting for the regression model and XGBoost for the classification model. Based on the outcomes, we can determine whether COLD-derived data provides sufficient information to estimate water frequency and accurately classify each pixel as either water or non-water.
\subsection{Determining Water Changes at Breakpoints in COLD-derived data}

Pixel data in COLD captures multiple coefficients for several stable periods. A breakpoint marks the transition between periods where the previous model no longer fits the new Landsat time series, requiring the model's coefficients to change to reflect spectral variations, as shown in Figure \ref{fig:general}. In this experiment, we input the COLD coefficients of pixels with breakpoints into a regression model to predict water frequency before and after the break. This analysis helped us identify whether the change was due to shifts in the water frequency across the stable periods surrounding the breakpoint, as illustrated in Figure \ref{fig:general}.

The three classes used in this experiment were determined by the absolute difference in the predicted water frequency index between periods before and after the break. To minimize computational error, the first class, labeled as 0, was assigned to cases where the absolute difference was close to zero or less than 0.25. Pixels with a difference greater than 0.25 were classified as an increasing water frequency class, whereas those with a difference smaller than –0.25 were classified as a decreasing water frequency class. 

\section{Results} 
This section evaluates classification and regression models to estimate water frequency and segment water and non-water bodies using COLD-derived data. Ground truth is derived from water indices from Landsat data. We apply machine learning models to estimate water frequency and classify pixels based on water frequency during stable COLD periods, using 7-fold cross-validation for unbiased results. The models are trained on six regions and tested on the remaining one to demonstrate robustness in detecting water bodies and trends. Visualizations highlight the results, showcasing COLD-derived data's potential for water segmentation and change detection.
\subsection{Water Frequency Prediction Using a Regression Model}

In this experiment, the input to our models consists of water frequency data across available years during the stable period of the COLD-derived dataset. We apply a gradient boosting regression model to estimate water frequency for each pixel during these stable periods, facilitating the identification of whether pixels represent water bodies or non-water bodies. For inference, data from one region is held out while the model is trained on the remaining six regions of interest.

The normalized mean squared error (NMSE) of our regression model is reported with an average value of 0.43 and a 95\% confidence interval of 0.12 to 0.75. This indicates that the model's predictive performance surpasses that of a baseline model, which would simply predict the mean of the actual values. The model demonstrates strong predictive power, as it captures a significant portion of the variance in the data. However, the upper bound of 0.75 suggests that the model exhibits considerably higher error in some cases, likely due to variations in environmental conditions across different regions.

We illustrate the model's ability to segment water bodies by applying a 0.25 threshold to the predicted water frequency. Pixels below this threshold are classified as non-water, and those above are classified as water bodies \citep{fengLongtermDenseLandsat2022}. Water body classification accuracy on the inference dataset is shown in Table \ref{tab:model_comparison_water_body_classification}. In the water body map derived from the regression model (Figure \ref{fig:water_bodies}, middle), darker blue pixels represent periods covered by the training data, while lighter blue pixels represent periods beyond the training scope. Both accurately depict water bodies, demonstrating the model’s robustness across time. Brown pixels indicate non-water bodies outside the training period, while dark brown pixels represent non-water within the training period.

\subsection{Assessing Water Frequency Prediction: Classification vs. Regression Models}

\begin{table}[htbp]
\centering
\caption{Cross-Validation Water Body Classification: Regression vs. Classification Model Inference Accuracy}
\begin{tabular}{|c|c|c|}
\hline
\textbf{}                            & \textbf{Regression}  & \textbf{Classification} \\ \hline
\textbf{Mean Overall Accuracy}             & 0.90 ± 0.06                        & 0.91 ± 0.07                            \\ \hline
\textbf{Water Frequency $\leq$ 0.25}       & 0.92 ± 0.07                        & 0.94 ± 0.09                            \\ \hline
\textbf{Water Frequency $>$ 0.25}          & 0.77 ± 0.11                        & 0.78 ± 0.10                            \\ \hline
\end{tabular}
\label{tab:model_comparison_water_body_classification}
\end{table}

In this experiment, we compared classification model with the regression model to delineate water frequency. For ground truth, we used a threshold of 0.25 to classify pixels as either water bodies or non-water bodies \citep{fengLongtermDenseLandsat2022}. Among the various models, we selected XGBoost as the best performer for classifying water bodies. Our classification result comes out as shown in Table \ref{tab:model_comparison_water_body_classification}. The model performs reliably for clear cases, such as low and high-frequency water bodies. 

\begin{figure}[htbp]
    \centering
    \begin{subfigure}{0.323\linewidth}
        \centering
        \includegraphics[width=\linewidth]{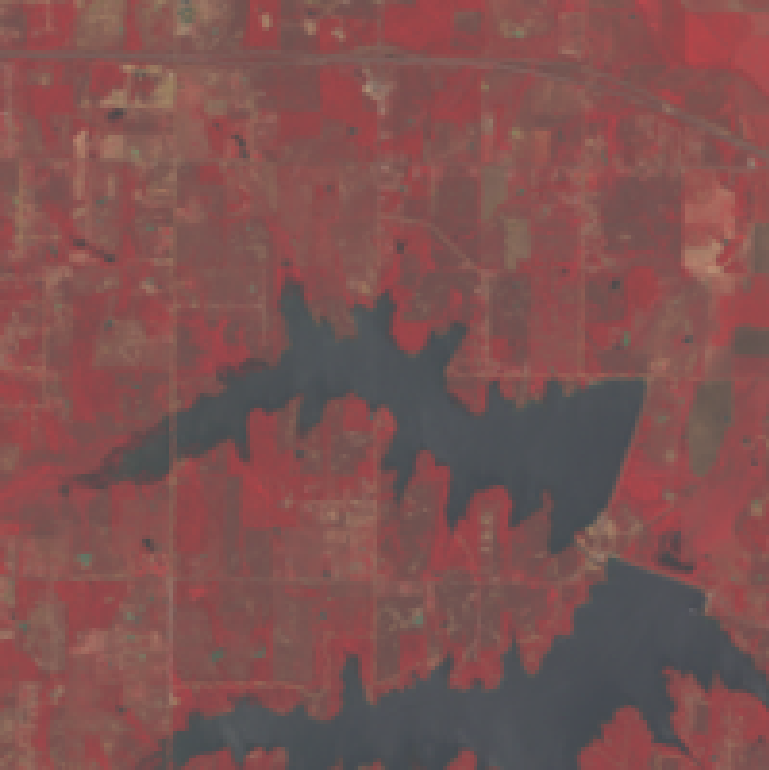}
        \label{fig:1_classifcation_inferred_map_a}
    \end{subfigure}
    \begin{subfigure}{0.327\linewidth}
        \centering
        \includegraphics[width=\linewidth]{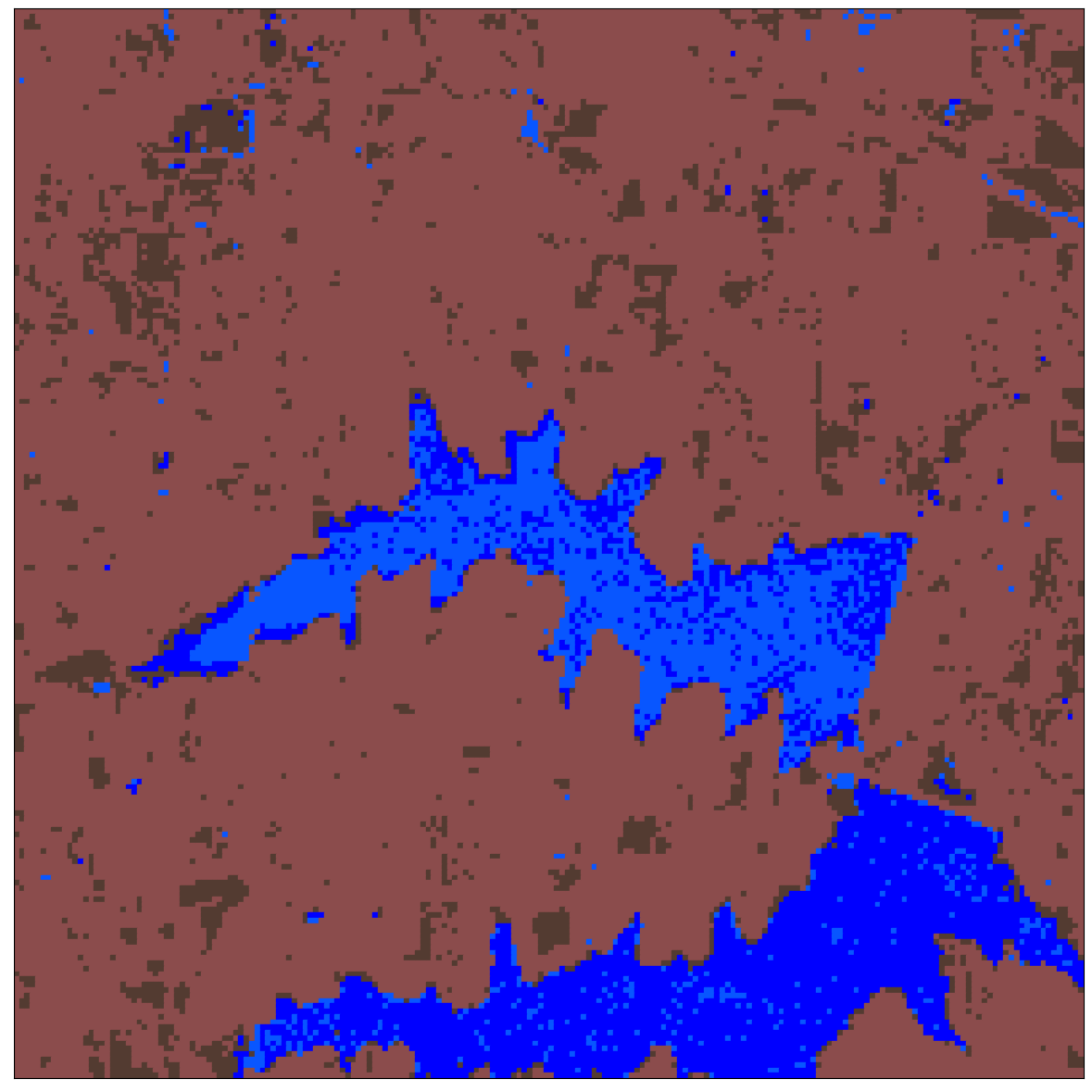}
        \label{fig:1_classifcation_inferred_map_b}
    \end{subfigure}
    \begin{subfigure}{0.327\linewidth}
        \centering
        \includegraphics[width=\linewidth]{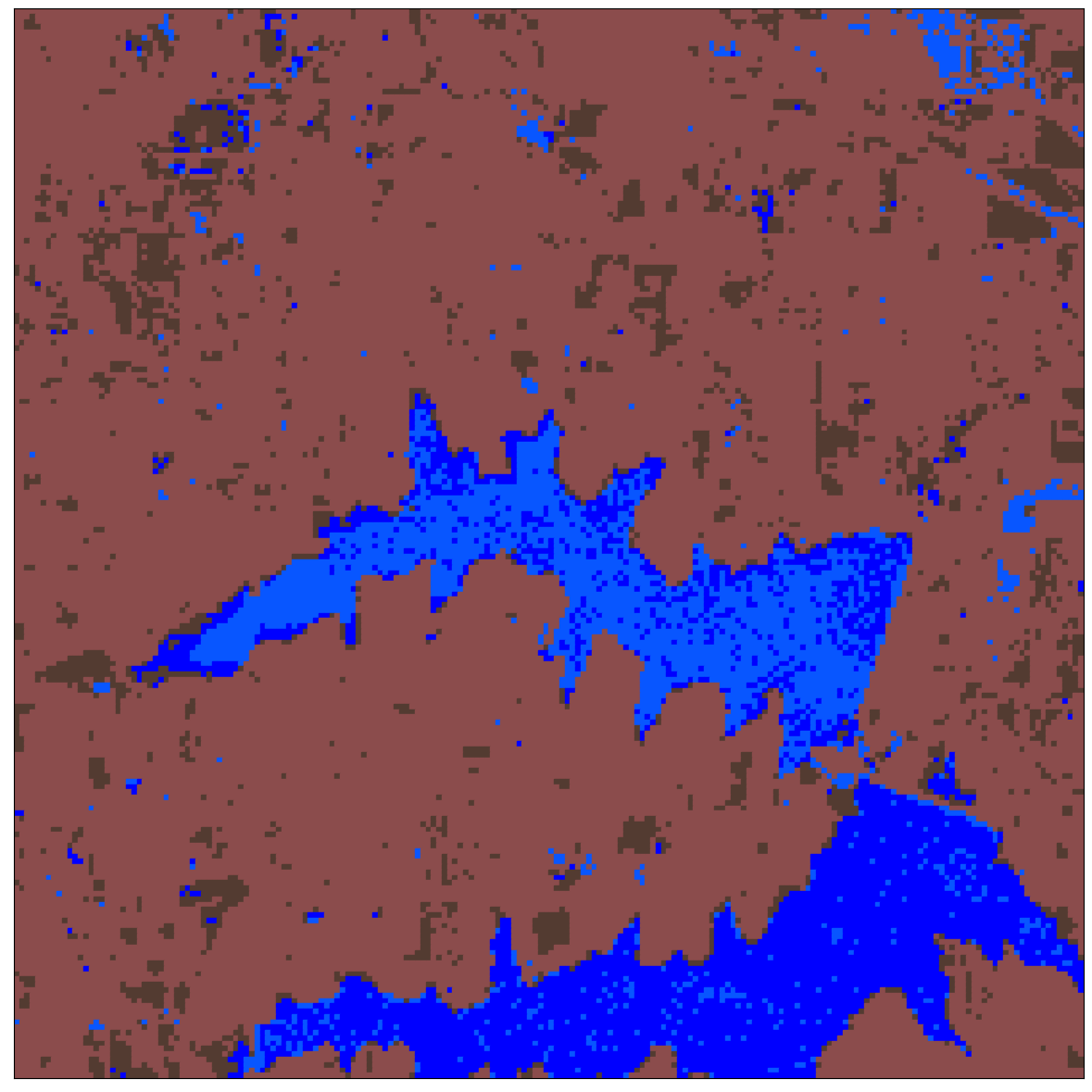}
        \label{fig:classification_model_1_compare_region_1_b}
    \end{subfigure}
    \caption{Visualization of water map based on the outcomes from the classification (middle) and regression model (right) in a region withs its Landsat image (left). Blue pixels indicate water bodies, while brown and dark brown pixels represent non-water bodies. Lighter-colored pixels represent the inference on pixels with stable periods beyond the timeframe of those used in the training process.}
    \label{fig:water_bodies}
\end{figure}

Additionally, our results indicate a slightly higher in performance when using the classification model as shown in Table \ref{tab:model_comparison_water_body_classification}. However, the confidence intervals overlap each other, indicating no significant statistical difference. The visualization between regression and classification map inference are shown in the Figure \ref{fig:water_bodies}. This comparison reveals that the regression model tends to over-segment the water bodies, as evidenced by the incorrect segmentation of the water bodies in the top-right region of the regression model.

\subsection{Determine water change using COLD-derived data}
This section analyzes pixel data with breakpoints between stable periods to determine whether water frequency trends increase or decrease. Pixels with breakpoints are filtered and used in a regression model to predict water frequency based on COLD coefficients before and after the breakpoint. Predicted frequency changes are classified as a decrease (less than -0.25), an increase (greater than 0.25), or unchanged (between -0.25 and 0.25). This method helps interpret significant changes at the breakpoint and how COLD coefficients respond. By analyzing these coefficients, we aim to predict the water frequency trend for each pixel at the breakpoint.

In addition, we also do experiments for determining water change using classification models instead of regression models. In this, we concatenate all COLD coefficients of before and after breakpoint together and use them to predict the differences between known water frequency before and after the breakpoint with the same range previously defined.

A comparison of regression and classification models shows that the regression model consistently outperforms the classification model across all metrics. The regression model's overall accuracy is $0.81 \pm 0.09$, compared to $0.77 \pm 0.14$ for the classification model, with a narrower confidence interval, indicating greater stability. In the moderate water frequency range ($-0.25 \leq WF \leq 0.25$), the regression model achieves $0.85 \pm 0.09$, outperforming the classification model’s $0.82 \pm 0.14$. For rare water frequency classes ($WF < -0.25$ and $WF > 0.25$), the regression model again performs better, with accuracies of $0.50 \pm 0.11$ and $0.42 \pm 0.14$, compared to $0.43 \pm 0.15$ and $0.31 \pm 0.12$ for the classification model. While both models struggle with these extreme cases, the regression model consistently shows stronger predictive power. It encounters challenges primarily when water frequency significantly changes, as these events are relatively rare. A method to address the imbalanced dataset, such as SMOTE \citep{chawlaSMOTESyntheticMinority2011}, has been applied to improve the model's performance by oversampling the minority class.

\begin{table}[htbp]
\centering
\caption{Cross-Validation Inference Performance for Water Change Detection}
\begin{tabular}{|c|c|c|}
\hline
\textbf{Metric}                            & \textbf{Regression Model}         & \textbf{Classification Model}    \\ \hline
\textbf{Overall Accuracy}                  & 0.81 ± 0.09                         & 0.77 ± 0.14                        \\ \hline
\textbf{-0.25 $\leq$ WF $\leq$ 0.25} & 0.85 ± 0.09                         & 0.82 ± 0.14                        \\ \hline
\textbf{WF $<$ -0.25}         & 0.50 ± 0.11                         & 0.43 ± 0.15                        \\ \hline
\textbf{WF $>$ 0.25}          & 0.42 ± 0.14                         & 0.31 ± 0.12                        \\ \hline
\end{tabular}
\label{tab:regression_classification_water_change}
\end{table}

\section{Discussion}
This study demonstrates the effectiveness of both regression and classification models in estimating water frequency and segmenting water bodies using COLD coefficient data \citep{zhuContinuousMonitoringLand2020}. The gradient boosting regression model achieved a normalized mean squared error (NMSE) of 0.43, indicating strong predictive capability, though certain regions showed higher errors, likely due to environmental variability. For water body segmentation, using a 0.25 water frequency threshold, the regression model reached 90\% accuracy, performing better in low water frequency areas (92\%) than in high frequency regions (77\%). A comparison of models showed similar performance, with the classification model slightly outperforming the regression model (91\% vs. 90\%), although the difference was not statistically significant. The regression model tended to over-segment water bodies in some areas. Despite limited training data, both models accurately predicted water frequency beyond the training period, as seen in Figure \ref{fig:water_bodies}. COLD-derived data detected land disturbances over time, with a correlation between its coefficients and water presence, suggesting discrepancies may be linked to water in affected regions. This insight enhances understanding of climate change impacts on water bodies.

In analyzing water frequency changes at breakpoints, the regression model outperformed the classification model, achieving an overall accuracy of 81\% compared to 77\%. The regression model was particularly effective in predicting moderate changes in water frequency but faced challenges in capturing extreme changes. This issue is likely due to the imbalanced training data, where instances of unchanged or moderate water frequency differences were more common than significant changes. Consequently, the model performed better in handling these more prevalent cases. Another potential factor is the variation in data distribution across different regions. Overall, the regression model consistently outperformed the classification model, demonstrating its ability to detect trends in water frequency at breakpoints. However, there is still room for improvement in detecting water trends, particularly in identifying increases or decreases in water frequency. Future work could expand the study by collecting additional data or generating realistic synthetic data for the minority class \citep{maryadaEfficientSyntheticData2022} to improve the model’s performance.

Our results are consistent with previous studies on land cover classification, such as \citep{zhuContinuousChangeDetection2014}, which employed the Continuous Change Detection and Classification (CCDC) algorithm—an earlier version of COLD—to classify land cover types using a classification model. In this study, we utilized a regression model to estimate water frequency, comparing it with a classification model to map water bodies and analyze trends based on data from the updated CCDC algorithm (COLD). Direct comparisons between our approach and previous work using the CCDC model are not feasible due to differences in data sources (CCDC-derived vs. COLD-derived data) and variations in landscape and data collection methods. The primary contribution of our research is the validation and extension of COLD's capacity to detect water bodies and identify regional trends, which supports water resource management.

We used the water index as the primary indicator of water bodies, serving as ground truth to evaluate correlations between water and non-water pixels in COLD-derived data. While water indices, widely used in past studies, they still contain errors. We chose these indices for their lower cost, using Landsat data. Future work could improve accuracy by incorporating natural images from Google Earth Engine, captured during the same periods as the COLD-derived data, to better establish water body ground truth.

\section{Conclusion}

This study demonstrates the effectiveness of the COLD algorithm for water frequency estimation and water body segmentation in aquatic ecosystems. Both regression and classification models applied to coefficients of COLD-derived data produce reliable results, with the regression model achieving a normalized mean squared error of 0.43 and 90\% accuracy in water mapping. Despite limited training data, our model accurately predicts water frequency beyond the training period. These results align with prior studies on land cover classification using the CCDC algorithm, extending COLD's applicability to diverse regions. The regression model also detects water frequency changes at breakpoints with 81\% accuracy, though it faces challenges with extreme changes due to imbalanced data. Future research could improve accuracy by incorporating higher-resolution indicators and natural imagery, strengthening its use in environmental monitoring and conservation.


\section*{Funding}

This work was self-supported.

\section*{Disclosure statement}
The authors report no conflict of interest.
\bibliographystyle{tfcad}
\bibliography{_main.bib}

\begin{thebibliography}{32}
\newcommand{\enquote}[1]{``#1''}
\providecommand{\natexlab}[1]{#1}
\providecommand{\url}[1]{\normalfont{#1}}
\providecommand{\urlprefix}{}

\bibitem[Bijeesh and Narasimhamurthy(2021)]{bijeeshGenericFrameworkChange2021}
Bijeesh, T.~V., and K.~N. Narasimhamurthy. 2021. ``A {{Generic Framework}} for {{Change Detection}} on {{Surface Water Bodies Using Landsat Time Series Data}}.'' In \emph{Computational {{Vision}} and {{Bio-Inspired Computing}}},  edited by S.~Smys, Jo{\~a}o Manuel R.~S. Tavares, Robert Bestak, and Fuqian Shi, Vol. 1318, 303--314. Singapore: Springer Singapore.

\bibitem[Cao et~al.(2024)]{caoWaterBodyExtraction2024}
Cao, Huidong, Yanbing Tian, Yanli Liu, and Ruihua Wang. 2024. ``Water Body Extraction from High Spatial Resolution Remote Sensing Images Based on Enhanced {{U-Net}} and Multi-Scale Information Fusion.'' \emph{Scientific Reports} 14 (1): 16132.

\bibitem[Chang et~al.(2024)]{changMultiScaleAttentionNetwork2024}
Chang, Jing, Xiaohui He, Panle Li, Ting Tian, Xijie Cheng, Mengjia Qiao, Tao Zhou, Beibei Zhang, Ziqian Chang, and Tingwei Fan. 2024. ``Multi-{{Scale Attention Network}} for {{Building Extraction}} from {{High-Resolution Remote Sensing Images}}.'' \emph{Sensors} 24 (3): 1010.

\bibitem[Chawla et~al.(2011)]{chawlaSMOTESyntheticMinority2011}
Chawla, N.~V., K.~W. Bowyer, L.~O. Hall, and W.~P. Kegelmeyer. 2011. ``{{SMOTE}}: {{Synthetic Minority Over-sampling Technique}}.'' Jun.

\bibitem[Chisadza et~al.(2022)]{chisadzaDetectingLandSurface2022}
Chisadza, Bright, Onalenna Gwate, France Ncube, Nkosinathi Moyo, and Phibion Chiwara. 2022. ``Detecting Land Surface Water Changes in the {{Upper Mzingwane}} Sub-Catchment Using Remotely Sensed Data.'' \emph{Journal of Water Supply: Research and Technology-Aqua} 71 (10): 1180--1196.

\bibitem[Duan et~al.(2021)]{duanRecentAdvancementRemote2021}
Duan, Weili, Shreedhar Maskey, Pedro L.~B. Chaffe, Pingping Luo, Bin He, Yiping Wu, and Jingming Hou. 2021. ``Recent {{Advancement}} in {{Remote Sensing Technology}} for {{Hydrology Analysis}} and {{Water Resources Management}}.'' \emph{Remote Sensing} 13 (6): 1097.

\bibitem[Dwyer et~al.(2018)]{dwyerAnalysisReadyData2018}
Dwyer, John~L., David~P. Roy, Brian Sauer, Calli~B. Jenkerson, Hankui~K. Zhang, and Leo Lymburner. 2018. ``Analysis {{Ready Data}}: {{Enabling Analysis}} of the {{Landsat Archive}}.'' \emph{Remote Sensing} 10 (9): 1363.

\bibitem[Erdem et~al.(2021)]{erdemEnsembleDeepLearning2021}
Erdem, Firat, Bulent Bayram, Tolga Bakirman, Onur~Can Bayrak, and Burak Akpinar. 2021. ``An Ensemble Deep Learning Based Shoreline Segmentation Approach ({{WaterNet}}) from {{Landsat}} 8 {{OLI}} Images.'' \emph{Advances in Space Research} 67 (3): 964--974.

\bibitem[Feng et~al.(2022)]{fengLongtermDenseLandsat2022}
Feng, Shuailong, Shuguang Liu, Guoyi Zhou, Cheng Gao, Dong Sheng, Wende Yan, Yiping Wu, et~al. 2022. ``Long-Term Dense {{Landsat}} Observations Reveal Detailed Waterbody Dynamics and Temporal Changes of the Size-Abundance Relationship.'' \emph{Journal of Hydrology: Regional Studies} 41: 101111.

\bibitem[Gharbia(2023)]{gharbiaDeepLearningAutomatic2023}
Gharbia, Reham. 2023. ``Deep {{Learning}} for {{Automatic Extraction}} of {{Water Bodies Using Satellite Imagery}}.'' \emph{Journal of the Indian Society of Remote Sensing} 51 (7): 1511--1521.

\bibitem[Gong et~al.(2020)]{gongApplicabilitySurfaceWater2020}
Gong, Wenfeng, Tiedong Liu, Yan Jiang, and Philip Stott. 2020. ``Applicability of the {{Surface Water Extraction Methods Based}} on {{China}}'s {{GF-2 HD Satellite}} in {{Ussuri River}}, {{Tonghe County}} of {{Northeast China}}.'' \emph{Nature Environment and Pollution Technology} 19 (4): 1537--1545.

\bibitem[Huang et~al.(2018)]{huangDetectingExtractingMonitoring2018}
Huang, Chang, Yun Chen, Shiqiang Zhang, and Jianping Wu. 2018. ``Detecting, {{Extracting}}, and {{Monitoring Surface Water From Space Using Optical Sensors}}: {{A Review}}.'' \emph{Reviews of Geophysics} 56 (2): 333--360.

\bibitem[Kirillov et~al.(2023)]{kirillovSegmentAnything2023c}
Kirillov, Alexander, Eric Mintun, Nikhila Ravi, Hanzi Mao, Chloe Rolland, Laura Gustafson, Tete Xiao, et~al. 2023. ``Segment {{Anything}}.'' Apr.

\bibitem[Liu et~al.(2021)]{liuInvestigationCapabilityMultitemporal2021}
Liu, Di, Zhixin Qi, Hui Zhang, Xia Li, Anthony Gar-on Yeh, and Jiao Wang. 2021. ``Investigation of the Capability of Multitemporal {{RADARSAT-2}} Fully Polarimetric {{SAR}} Images for Land Cover Classification: A Case of {{Panyu}}, {{Guangdong}} Province.'' \emph{European Journal of Remote Sensing} 54 (1): 338--350.

\bibitem[Liu, Liu, and Hu(2024)]{liuNovelDeepLearning2024}
Liu, Min, Jiangping Liu, and Hua Hu. 2024. ``A {{Novel Deep Learning Network Model}} for {{Extracting Lake Water Bodies}} from {{Remote Sensing Images}}.'' \emph{Applied Sciences} 14 (4): 1344.

\bibitem[Mahua, Kasim, and Pasisingi(2023)]{mahuaANALYSISCOASTLINECHANGES2023}
Mahua, Musdalifah, Faizal Kasim, and Nuralim Pasisingi. 2023. ``{{ANALYSIS OF COASTLINE CHANGES IN GORONTALO CITY USING REMOTE SENSING TECHNOLOGY}}.'' \emph{Jurnal Ilmu Kelautan SPERMONDE} 39--46.

\bibitem[Maryada et~al.(2022)]{maryadaEfficientSyntheticData2022}
Maryada, Sai~Kiran, William~Lee Booker, Gopichandh Danala, Meredith Jones, Sanjana Mudduluru, Huong Pham, Dean~F. Hougen, and Bin Zheng. 2022. ``An {{Efficient Synthetic Data Generation Algorithm}} to {{Improve Efficacy}} of {{Deep Learning Models}} of {{Medical Images}}.'' Dec.

\bibitem[Melichar et~al.(2023)]{melicharRandomForestClassification2023}
Melichar, Madeline, Kamel Didan, Armando {Barreto-Mu{\~n}oz}, Jennifer~N. Duberstein, Eduardo Jim{\'e}nez~Hern{\'a}ndez, Theresa Crimmins, Haiquan Li, Myles Traphagen, Kathryn~A. Thomas, and Pamela~L. Nagler. 2023. ``Random {{Forest Classification}} of {{Multitemporal Landsat}} 8 {{Spectral Data}} and {{Phenology Metrics}} for {{Land Cover Mapping}} in the {{Sonoran}} and {{Mojave Deserts}}.'' \emph{Remote Sensing} 15 (5): 1266.

\bibitem[Minh~Trinh(2024)]{minhtrinhApplicationMNDWIIndex2024}
Minh~Trinh, Ngoc. 2024. ``Application of {{MNDWI}} Index for Flood Damage Area Calculation in {{Lam}} River Basin Using Google Earth Engine Platform.'' \emph{Journal of Hydro-meteorology} 8 (19): 1--11.

\bibitem[Mullen et~al.(2023)]{mullenUsingHighResolutionSatellite2023}
Mullen, Andrew~L., Jennifer~D. Watts, Brendan~M. Rogers, Mark~L. Carroll, Clayton~D. Elder, Jonas Noomah, Zachary Williams, et~al. 2023. ``Using {{High-Resolution Satellite Imagery}} and {{Deep Learning}} to {{Track Dynamic Seasonality}} in {{Small Water Bodies}}.'' \emph{Geophysical Research Letters} 50 (7): e2022GL102327.

\bibitem[Nasir et~al.(2023)]{nasirDeepLearningDetection2023}
Nasir, Nida, Afreen Kansal, Omar Alshaltone, Feras Barneih, Abdallah Shanableh, Mohammad {Al-Shabi}, and Ahmed Al~Shammaa. 2023. ``Deep Learning Detection of Types of Water-Bodies Using Optical Variables and Ensembling.'' \emph{Intelligent Systems with Applications} 18: 200222.

\bibitem[Ozdemir et~al.(2024)]{ozdemirExtractionWaterBodies2024}
Ozdemir, Samed, Zeynep Akbulut, Fevzi Karsli, and Taskin Kavzoglu. 2024. ``Extraction of {{Water Bodies}} from {{High-Resolution Aerial}} and {{Satellite Images Using Visual Foundation Models}}.'' \emph{Sustainability} 16 (7): 2995.

\bibitem[Radford et~al.(2021)]{radfordLearningTransferableVisual2021a}
Radford, Alec, Jong~Wook Kim, Chris Hallacy, Aditya Ramesh, Gabriel Goh, Sandhini Agarwal, Girish Sastry, et~al. 2021. ``Learning {{Transferable Visual Models From Natural Language Supervision}}.'' Feb.

\bibitem[Shetty et~al.(2021)]{shettyAssessingEffectTraining2021}
Shetty, Shobitha, Prasun~Kumar Gupta, Mariana Belgiu, and S.~K. Srivastav. 2021. ``Assessing the {{Effect}} of {{Training Sampling Design}} on the {{Performance}} of {{Machine Learning Classifiers}} for {{Land Cover Mapping Using Multi-Temporal Remote Sensing Data}} and {{Google Earth Engine}}.'' \emph{Remote Sensing} 13 (8): 1433.

\bibitem[Wang and Yao(2024)]{wangMountainVegetationClassification2024}
Wang, Baoguo, and Yonghui Yao. 2024. ``Mountain {{Vegetation Classification Method Based}} on {{Multi-Channel Semantic Segmentation Model}}.'' \emph{Remote Sensing} 16 (2): 256.

\bibitem[Wang et~al.(2023)]{wangSAMRSScalingupRemote2023}
Wang, Di, Jing Zhang, Bo~Du, Minqiang Xu, Lin Liu, Dacheng Tao, and Liangpei Zhang. 2023. ``{{SAMRS}}: {{Scaling-up Remote Sensing Segmentation Dataset}} with {{Segment Anything Model}}.'' Oct.

\bibitem[Wang et~al.(2020)]{wangTrackingAnnualChanges2020}
Wang, Xinxin, Xiangming Xiao, Zhenhua Zou, Bangqian Chen, Jun Ma, Jinwei Dong, Russell~B. Doughty, et~al. 2020. ``Tracking Annual Changes of Coastal Tidal Flats in {{China}} during 1986-2016 through Analyses of {{Landsat}} Images with {{Google Earth Engine}}.'' \emph{Remote sensing of environment} 238: 110987.

\bibitem[Xu(2006)]{xuModificationNormalisedDifference2006}
Xu, Hanqiu. 2006. ``Modification of Normalised Difference Water Index ({{NDWI}}) to Enhance Open Water Features in Remotely Sensed Imagery.'' \emph{International Journal of Remote Sensing} 27 (14): 3025--3033.

\bibitem[Yue et~al.(2023)]{yueFullyAutomaticHighaccuracy2023}
Yue, Linwei, Baoguang Li, Shuang Zhu, Qiangqiang Yuan, and Huanfeng Shen. 2023. ``A Fully Automatic and High-Accuracy Surface Water Mapping Framework on {{Google Earth Engine}} Using {{Landsat}} Time-Series.'' \emph{International Journal of Digital Earth} 16 (1): 210--233.

\bibitem[Yulianto et~al.(2022)]{yuliantoEvaluationThresholdImproved2022}
Yulianto, Fajar, Dony Kushardono, Syarif Budhiman, Gatot Nugroho, Galdita~Aruba Chulafak, Esthi~Kurnia Dewi, and Anjar~Ilham Pambudi. 2022. ``Evaluation of the {{Threshold}} for an {{Improved Surface Water Extraction Index Using Optical Remote Sensing Data}}.'' \emph{The Scientific World Journal} 2022: 4894929.

\bibitem[Zhu and Woodcock(2014)]{zhuContinuousChangeDetection2014}
Zhu, Zhe, and Curtis~E. Woodcock. 2014. ``Continuous Change Detection and Classification of Land Cover Using All Available {{Landsat}} Data.'' \emph{Remote Sensing of Environment} 144: 152--171.

\bibitem[Zhu et~al.(2020)]{zhuContinuousMonitoringLand2020}
Zhu, Zhe, Junxue Zhang, Zhiqiang Yang, Amal~H. Aljaddani, Warren~B. Cohen, Shi Qiu, and Congliang Zhou. 2020. ``Continuous Monitoring of Land Disturbance Based on {{Landsat}} Time Series.'' \emph{Remote Sensing of Environment} 238: 111116.

\end{thebibliography}

\end{document}